\documentclass[lettersize,journal]{IEEEtran}
\usepackage{amsmath,amsfonts}
\usepackage{algorithmic}
\usepackage{algorithm}
\usepackage{array}
\usepackage[caption=false,font=normalsize,labelfont=sf,textfont=sf]{subfig}
\usepackage{textcomp}
\usepackage{stfloats}
\usepackage{url}
\usepackage{verbatim}
\usepackage{graphicx}
\usepackage{cite}
\usepackage{tabularx}

\usepackage[absolute,showboxes]{textpos}

\TPMargin{5pt}

\newcommand{\copyrightstatement}{
    \begin{textblock}{0.84}(0.08,0.96)
         \noindent
         \footnotesize
         This work has been submitted to the IEEE for possible publication. Copyright may be transferred without notice, after which this version may no longer be accessible.
    \end{textblock}
}

\begin{document}
\copyrightstatement
% \submittednotice
% \lipsum[1-10]

\title{Integrating Secondary Structures Information into Triangular Spatial Relationships (TSR) for Advanced Protein Classification}

\author{Poorya Khajouie, Titli Sarkar, Krishna Rauniyar, Li Chen, Wu Xu, Vijay Raghavan ~\IEEEmembership{Member,~IEEE,}
        % <-this % stops a space
\thanks{This study is supported by NIH NIGMS (1R15GM144944-01) and the Louisiana Board of Regents (LEQSF(2023-24)-RD-D-05).}% <-this % stops a space
\thanks{Poorya Khajouie, Krishna Rauniyar, Titli Sarkar, Li Chen, and Vijay Raghavan are with The School of Computing and
Informatic, University of Louisiana at Lafayette, LA 70504, USA (email: poorya.khajouie1@louisiana.edu, titli010203@gmail.com, krishna.rauniyar1@louisiana.edu, li.chen@louisiana.edu,
vijay.raghavan@louisiana.edu).}
\thanks{Wu Xu is with the Department of Chemistry, University of Louisiana at Lafayette, P.O. Box 44370, Lafayette, LA 70504, USA (email: wu.xu@louisiana.edu).}

\thanks{For reproducibility and further research, the code used in this study is accessible at https://github.com/pooryakhajouie/SSE-TSR.\\ \textit{(Corresponding Author: Vijay Raghavan, Co-corresponding Author: Wu Xu)}}
}
\maketitle

\begin{abstract}

Protein structures represent the key to deciphering biological functions. The more detailed form of similarity among these proteins is sometimes overlooked by the conventional structural comparison methods. In contrast, further advanced methods, such as Triangular Spatial Relationship (TSR), have been demonstrated to make finer differentiations. Still, the classical implementation of TSR does not provide for the integration of secondary structure information, which is important for a more detailed understanding of the folding pattern of a protein. To overcome these limitations, we developed the SSE-TSR approach. The proposed method integrates secondary structure elements (SSEs) into TSR-based protein representations. This allows an enriched representation of protein structures by considering 18 different combinations of helix, strand, and coil arrangements. Our results show that using SSEs improves the accuracy and reliability of protein classification to varying degrees. We worked with two large protein datasets of 9.2K and 7.8K samples, respectively. We applied the SSE-TSR approach and used a neural network model for classification. Interestingly, introducing SSEs improved performance statistics for Dataset 1, with accuracy moving from 96.0\% to 98.3\%. For Dataset 2, where the performance statistics were already good, further small improvements were found with the introduction of SSE, giving an accuracy of 99.5\% compared to 99.4\%. These results show that SSE integration can dramatically improve TSR key discrimination, with significant benefits in datasets with low initial accuracies and only incremental gains in those with high baseline performance. Thus, SSE-TSR is a powerful bioinformatics tool that improves protein classification and understanding of protein function and interaction.
\end{abstract}


\section{Introduction}
\IEEEPARstart{P}{rotein} structural analysis is critical for understanding their biological functions and interactions. As complicated macromolecules, proteins show a wide spectrum of structural conformations, which makes classification and comparison of them both vital and difficult. Conventional techniques for structural comparison usually rely on sequence alignment and 3D structural superposition, which can be computationally demanding and occasionally inadequate for identifying subtle similarities across several protein structures \cite{smith1981identification, jumper2021highly}.

Understanding 3D structures, not just 1D connections, gives us a better picture of how proteins work. Many algorithms have been created to compare structures, such as the maximal common subgraph detection algorithm \cite{bron1973algorithm}, the Ullmann subgraph isomorphism algorithm \cite{ullmann1976algorithm}, the geometric hashing algorithm \cite{nussinov1991efficient}, the Monte Carlo algorithm \cite{holm1993protein}, the Combinatorial Extension (CE) algorithm \cite{shindyalov1998protein}, and Comparative Structural Alignment (CSA) algorithm \cite{wohlers2012csa}. Advanced techniques such as genetic algorithms \cite{szustakowski2000protein}, Dictionary of Secondary Structure of Proteins (DSSP) \cite{kabsch1983dictionary}, and amino acid networks (AAN) \cite{alves2007inferring, bartoli2008effect}, including C$\alpha$ network (CAN), atom distance network (ADN), and interaction selective network (ISN) \cite{konno2019quantitative}, are also widely used. Dynamic programming algorithms are used for distance-based \cite{blundell1988knowledge, taylor1989protein, lackner2000prosup} and secondary structure-based \cite{taylor1989protein, yang2000integrated} comparisons. Results from these methods can vary greatly due to computational effort, inconsistent outputs, sensitivity to structural variations, and alignment dependency \cite{mayr2007comparative, kolodny2005comprehensive}. Research indicates that combining approaches can enhance structural relationships and overcome restrictions \cite{wohlers2012csa, kolodny2005comprehensive}.
\IEEEpubidadjcol

To address these issues, the notion of Triangular Spatial Relationship (TSR) was introduced as a novel way for modeling protein structures \cite{kondra2021development, milon2024development}. TSR creates triangles with the C$\alpha$ atoms of proteins as vertices. Each triangle has a unique integer "key" determined based on geometric features such as length, angle, and vertex labels. This method ensures the consistent assignment of keys to identical TSRs across proteins, hence allowing a simpler but exact representation of protein structures. Using a dataset of 101 protein structures, we previously evaluated our TSR-based method against three well-known techniques: DALI, CE, and TM-align. The investigation revealed that although all methods effectively grouped proteins, the TSR-based method exhibited a greater capability to detect more subtle substructural variations \cite{kondra2021development}. This was evident from its higher Adjusted Rand Index (ARI) and bigger percentage of weighted distance. The results demonstrated that the TSR-based technique is susceptible to substructural similarities, establishing a solid basis for future enhancements. Motivated by these results, we aimed to enhance our method by integrating secondary structure information into the representation of protein structures. Focusing on the spatial arrangement of essential structural elements allows for an efficient and alignment-free technique of comparing and classifying proteins based on the number of common keys \cite{kondra2022study}.

Alpha helices, beta sheets, and coils are important components of protein structure. They have a significant impact on the overall 3D arrangement and functional characteristics of proteins. The hydrogen bonds stabilize these secondary structural components and play a crucial role in the folding and stability of proteins \cite{pauling1951configurations}. Classifying protein features into several categories based on secondary structure information offers useful insights into the structural and functional variability of proteins \cite{chothia1984principles}. 

Three amino acids and triangle geometry are represented by TSR keys. These keys are considered high-dimensional features for representing protein structures. However, the current version of the method does not account for secondary structures. Secondary structures underpin the architectural organization in proteins and are therefore key to understanding their 3D folding patterns \cite{pauling1951configurations, kabsch1983dictionary}. Pauling and Corey first defined two main secondary structural elements (SSEs), alpha helix and beta sheet, based on the intra-backbone hydrogen bond patterns in proteins \cite{pauling1951configurations}. Now, these specific motifs are almost ubiquitous across known structures \cite{konagurthu2012minimum}. Residues in known protein structures are approximately 30\% in helices, 20\% in strands, and 50\% in neither. SSEs are typically grouped into three larger classes: helix, strand, and coil \cite{labesse1997p}. Secondary structures have been extensively employed in structure visualization \cite{humphrey1996vmd}, classification \cite{sillitoe2015cath}, comparison \cite{madej1995threading}, and prediction \cite{sidi2020redundancy, klausen2019netsurfp}. Protein secondary structures can be roughly divided into three classes: mainly alpha ($\alpha$), mainly beta ($\beta$), and alpha-beta ($\alpha\beta$).

Our improved technique classifies the secondary structure type associated with a TSR key into 18 discrete categories, allocating each key to one of these categories to generate a more intricate and informative representation of protein structures. This methodology encompasses the geometric attributes of proteins and incorporates their secondary structural traits, offering a full and precise presentation for structural examination.

The necessity for sophisticated tools in protein structure prediction has become increasingly evident, as conventional methods, despite their effectiveness, are frequently constrained by their computational complexity and inability to capture intricate structural relationships. A transformative approach in this field has emerged: deep learning, a subset of machine learning, has the potential to substantially improve the accuracy and efficiency of protein structure prediction. Sophisticated models that are capable of learning hierarchical representations of protein sequences and structures have been developed as a result of recent advancements in deep learning \cite{jumper2021highly, senior2020improved}. This has provided a more profound understanding of protein's functional and interactive roles.

One of the most important developments in this field is AlphaFold, created by DeepMind, which, in some circumstances \cite{jumper2021highly}, shows amazing accuracy in predicting protein structures, so surpassing experimental approaches. Because of AlphaFold's success, many new deep learning models and methods have been developed to improve protein structure prediction. These models capture complex patterns in protein sequences utilizing large-scale datasets and sophisticated architectures, including convolutional neural networks (CNNs) and transformers \cite{alquraishi2019end, xu2021improved}. By combining deep learning with other computational methods, prediction capabilities have been further refined, enabling the analysis of large protein datasets with unparalleled speed and accuracy \cite{yang2020improved, torrisi2020deep, min2017deep}.

Aside from AlphaFold, several other deep learning frameworks have made significant advancements in the field. One instance is when RosettaFold enhances the precision of predictions \cite{baek2021accurate} by integrating deep learning with Rosetta's traditional physics-based modeling. Deep learning has played a crucial role in predicting protein-protein interactions, enabling a better understanding of complex biological processes and the creation of advanced therapies \cite{townshend2019end, gainza2020deciphering}. The quick growth of deep learning methods and their use in protein structure prediction highlights the need for greater research and development of reliable, scalable, and accurate bioinformatics tools. Recent research has also combined Bayesian optimization with protein structure prediction to simplify peptide design, illustrating the potential for innovative computational biology approaches.~\cite{manshour2023integrating, rao2021msa,lin2022language, jumper2021highly,senior2020improved,alquraishi2019end}. Furthermore, the adaptability of computational methods in addressing various bioinformatics challenges has been underscored by implementing machine learning techniques to forecast non-CG DNA methylation patterns~ \cite{sereshki2023prediction}.

This work presents a novel approach to generate TSR-based keys along with secondary structure information, and shows how to use it to classify proteins from two different datasets. Using the C$\alpha$ atoms of proteins as vertices, our method creates triangles and then assigns a unique key to each triangle using a rule-based formula comprising lengths, angles, and vertex labels. The 18 secondary structure groups are then used to categorize the TSR keys. This guarantees that TSR keys and the types of their secondary structures are consistently assigned across different proteins. To the best of our knowledge, this is the first time both 3D structure features and secondary structure features are explicitly incorporated into the representation of proteins. As a result, each protein can have a robust, alignment-free structural representation as a matrix of integers \cite{sarkar2023introducing}. We used our TSR-based approach to produce keys for two large protein datasets, adding secondary structure information to improve the structural representation. In addition to utilizing these improved representations, we implemented a deep neural network architecture that is compatible with 3D inputs for protein classification. This design can achieve significantly higher accuracy compared to classifications that did not consider secondary structure information. Our findings show how the SSE-TSR approach can enhance protein classification and structural analysis, providing a promising tool for bioinformatics research.
The subsequent sections elaborate on the methods employed, present our findings, and deliberate on the consequences of our discoveries concerning protein structure analysis and bioinformatics.

\section{Methodology}

\subsection{Datasets}

In this work, we used two distinct protein datasets categorized based on two different criteria: topology and function. The first dataset comprises three classes arranged according to protein topology: alpha (3392 samples, 33.4\%), alpha\_beta (3451 samples, 34.6\%), and beta (2430 samples, 24.4\%). This classification is based on the structural arrangement of the elements of the secondary structure of the proteins. Comprising five classes—Receptor (1436 samples, 14.6\%), Phosphatase (431 samples, 4.4\%), Kinase (2502 samples, 25.4\%), Isomerase (371 samples, 3.7\%), and Protease (3067 samples, 31.1\%), the second dataset is functionally categorized. This functional classification underscores the several roles these proteins play in biological processes. Unlike the topology-based dataset, which is rather balanced, the function-based dataset exhibits a more noticeable class imbalance, with protease and kinase classes dominant and isomerase and phosphatase classes underrepresented. Particularly when assessing how secondary structure information influences protein classification across datasets with different structural and functional categories, this distribution enables considerable in-depth evaluation of the SSE-TSR method.

\subsection{Key Generation}

\begin{figure*}[t!]
\centering
\includegraphics[trim=0 155 0 125, clip, width=\textwidth]{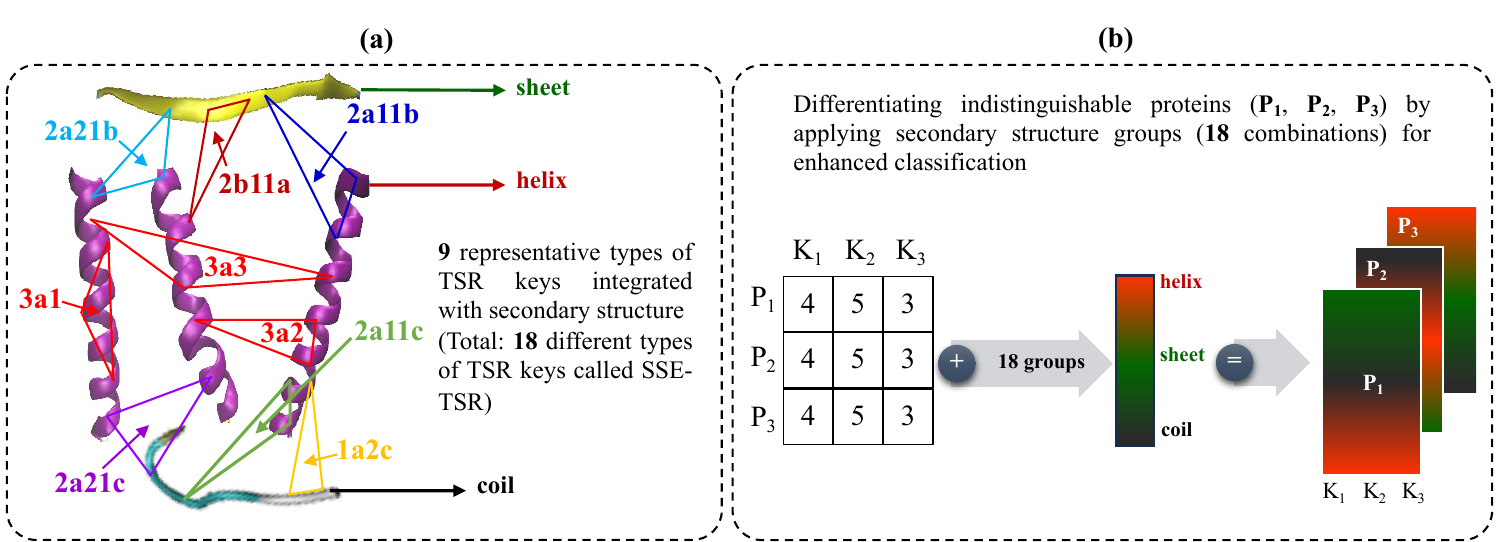}
\caption{Illustration highlighting the importance of integrating secondary structure elements with TSR keys to differentiate between TSR keys and proteins that were otherwise indistinguishable. The left panel displays 9 representative TSR keys associated with different SSE types (helices, sheets, coils), forming a total of 18 distinct SSE-TSR types. Proteins P1, P2, and P3, which at first glance seem to be identical based only on TSR keys, can be efficiently distinguished by classifying their keys into eighteen distinct combinations of secondary structure groups, as shown in the right panel. The proteins in this instance are identical, and the table displays the frequency of each key across proteins. The incorporation of secondary structural elements allows for the precise categorization of these proteins into different groups.}
\label{fig:sse-tsr}
\end{figure*}

C$\alpha$ atoms were extracted from the PDB (Protein Data Bank) file of each protein, forming the basis for computing all possible triangles' lengths and angles. To uniquely identify each C$\alpha$ of 20 amino acids, integers ranging from 4 to 23 were assigned. These integers were then converted into $l_{i1}$, $l_{i2}$, and $l_{i3}$, denoting the vertices of triangle $i$, employing a rule-based label-determination method \cite{guru2001triangular}. 

Upon determining $l_{i1}$, $l_{i2}$, and $l_{i3}$ for each triangle $i$, we calculated $\theta_{1}$ using Equation \ref{key_gen} and derived $\theta_{\Delta}$ based on the values of $\theta_{1}$.

\begin{equation}
\label{key_gen}
    \theta_{1} = cos^{-1}\Bigg(\Big(d_{13}^{\;2}-\big(\frac{d_{12}}{2}\big)^{2}-d_{3}^{\;2}\Big)/\Big(2\times\big(\frac{d_{12}}{2}\big)\times d_{3}\Big)\Bigg)
\end{equation}
\vspace{5pt}
\begin{align*}
\theta_{\Delta} =
    \begin{cases}
      \theta_{1} & \text{if $\theta \le 90^{\circ}$}\\
      180^{\circ} - \theta_{1} & \text{otherwise}
    \end{cases}
\end{align*}

where:
\begin{itemize}
    \item {$d_{13}$ is the distance between $l_{i1}$ and $l_{i3}$ for triangle $i$;}
    \item {$d_{12}$ is the distance between $l_{i1}$ and $l_{i2}$ for triangle $i$;}
    \item {$d_{3}$ is the distance between $l_{i3}$ and the midpoint of $l_{i1}$ and $l_{i2}$ for triangle $i$.}
\end{itemize}

The three side lengths of the triangle ($l_{i1}$, $l_{i2}$, and $l_{i3}$), an angle $\theta_{\Delta}$, and the longest distance, MaxDist ($D$), all contribute to determining the key value.

Theta ($\theta_{\Delta}$) is obtained from another angle, $\theta_{1}$. Theta is the angle, smaller than 90 degrees, produced between the line connecting the midpoint of the longest edge of the triangle and the vertex where the other two edges meet, and half of the longest edge. Theta essentially expresses the degree to which the triangle deviates from a right triangle.

MaxDist ($D$) is associated with the length of the triangle's longest edge. It functions as a scaling factor to guarantee that triangles with similar shapes but different sizes are assigned unique key values.

Upon calculating the labels $l_{i1}$, $l_{i2}$, $l_{i3}$, the maximum distance $D$, and $\theta_{\Delta}$, we utilize Equation \ref{key_cal} to compute the key value for each triangle.
\vspace{5pt}
\begin{multline}
\label{key_cal}
K_{3D} = \theta_{T}d_{T}(l_{i1}-1)m^{2} + \theta_{T}d_{T}(l_{i2}-1)m \\
+ \theta_{T}d_{T}(l_{i3}-1) + \theta_{T}(d-1) + (\theta-1)
\end{multline}
\vspace{5pt}
where:
\begin{itemize}
    \item {$m$ is the total number of distinct labels;}
    \item {$\theta$ is the bin value for the class in which $\theta_{\Delta}$, the angle representative, falls (using the Adaptive Unsupervised Iterative Discretization algorithm);}
    \item {$\theta_{T}$ is the total number of distinct discretization levels (or bins) for the angle representative;}
    \item {$d_T$ is the total number of distinct discretization levels (or bins) for the length representative.}
\end{itemize}

Without MaxDist, two analogous triangles of different sizes would share the same key value \cite{kondra2021development}.

\subsection{Incorporating Secondary Structure Information into TSR-Based Protein Structure Representation}

In protein structure representation, TSR (Triangle-Spatial Relationship) keys have emerged as a potent tool for encoding the geometric relationships of amino acids within triangles. However, the original methodology lacked integration with secondary structure information, a critical aspect for comprehending protein folding patterns. Secondary structures, which are classified as $\alpha$ helices, $\beta$ sheets, and coils, are essential components of protein architecture, providing valuable information regarding their three-dimensional organization \cite{konagurthu2012minimum, lesk1982computer}.

In order to fill this deficiency, we have created the SSE-TSR technique, which integrates secondary structure elements (SSEs) into TSR-based representations. SSEs, characterized by Pauling and Corey as $\alpha$ helices and $\beta$ sheets, constitute a substantial portion of known protein structures \cite{pauling1951structure}. Leveraging data from the Protein Data Bank (PDB), where approximately 30\% of residues reside in helices, 20\% in strands, and 50\% in coils \cite{berman2000protein}, the SSE-TSR method refines TSR keys by categorizing them based on the combination of secondary structure information of each vertex.

\subsection{Enhancing Protein Structure Representation with SSE-TSR Algorithm}

The SSE-TSR algorithm improves the representation of protein structure by integrating secondary structure information into TSR keys. The SSE-TSR approach accurately defines protein topologies through considering 18 unique combinations of helix, strand, and coil configurations. This algorithmic progress enables the differentiation of $\alpha\beta$ class from $\alpha$ or $\beta$ classes, as well as subclasses within $\alpha$, $\beta$, and $\alpha\beta$ structural patterns \cite{chou1978empirical, chou1974prediction}.

In order to ascertain the secondary structure type of each triangle that makes up a TSR key, we utilized a set of classification rules. The classification of triplets according to secondary structural elements is as follows:

\begin{enumerate}
\item{3a1: All three vertices from the same helix}
\item{3b1: All three vertices from the same sheet}
\item{3c: All vertices from coils}
\item{1a1b1c: One helix, one sheet, one coil}
\item{3a3: Three vertices from three different helices}
\item{3b3: Three vertices from three different sheets}
\item{3a2: Two vertices from the same helix, one vertex from another helix}
\item{2a11b: Two vertices from the same helix, one vertex from a sheet}
\item{2a11c: Two vertices from the same helix, one vertex from a coil}
\item{3b2: Two vertices from the same sheet, one vertex from another sheet}
\item{2b11a: Two vertices from the same sheet, one vertex from a helix}
\item{2b11c: Two vertices from the same sheet, one vertex from a coil}
\item{1a2c: One vertex from a helix, two vertices from coils}
\item{1b2c: One vertex from a sheet, two vertices from coils}
\item{2a21b: Two vertices from different helices, one vertex from a sheet}
\item{2a21c: Two vertices from different helices, one vertex from a coil}
\item{2b21a: Two vertices from different sheets, one vertex from a helix}
\item{2b21c: Two vertices from different sheets, one vertex from a coil}
\end{enumerate}

\begin{figure}[h!]
\centering
\includegraphics[trim=0 87 0 90, clip, width=3.5in]{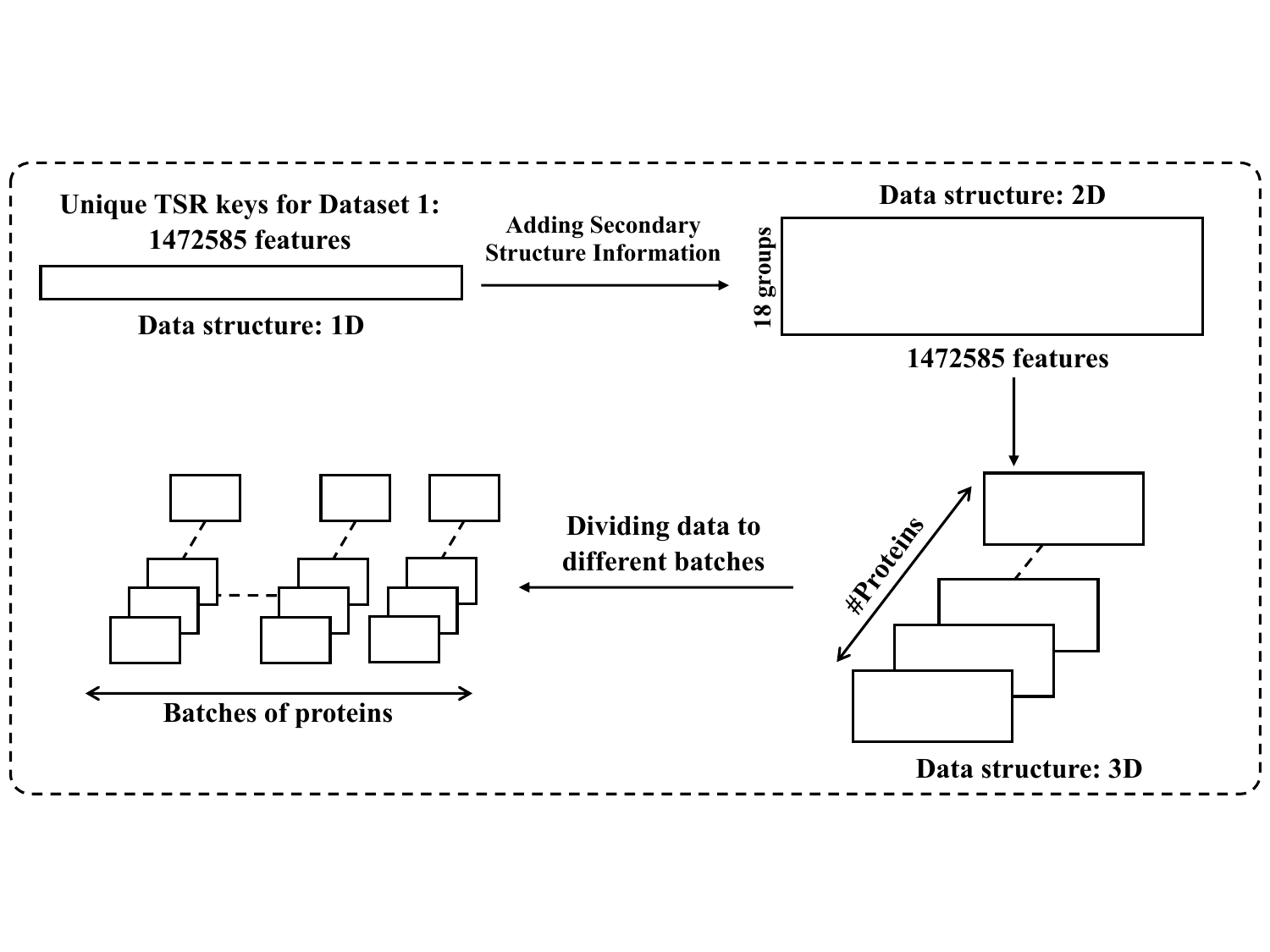}
\caption{Data structure transformation: (Top) Unique TSR keys forming a one-dimensional feature set. (Right) Addition of secondary structure information, creating a two-dimensional matrix. (Bottom) Batching the 2D matrices into a three-dimensional structure for efficient handling and processing.}
\label{fig:data_struct}
\end{figure}

The SSE-TSR algorithm converts the traditional 1D TSR key data structure into a 3D representation by assigning triangles to precise spatial positions. The exhibited structural change, as shown in Figure \ref{fig:data_struct}, not only improves visualization but also simplifies the investigation of similarities and differences in protein structure.

By integrating secondary structure elements into TSR keys, we introduced a critical enhancement in protein structure analysis. As depicted in Figure \ref{fig:sse-tsr}a, the TSR keys are combined with representative secondary structure types, forming a total of eighteen unique SSE-TSR combinations.  This integration greatly enhances the capacity to discriminate between various protein motifs by giving each TSR key extra contextual information on the protein's secondary structure. Figure \ref{fig:sse-tsr}b clearly illustrates how the inclusion of secondary structures enhances the ability to discriminate proteins, such as P1, P2, and P3, that would otherwise be indistinguishable when depending just on TSR keys.  The enhanced feature set not only increases the granularity and accuracy of the classification but also supports the core motivation of our research—improving the discriminative power of TSR keys by incorporating additional structural information to differentiate between protein structures more effectively.

\subsection{Data Preparation}

After creating key files together with their corresponding secondary structure information, our next task is to organize the data for the classification model's application. We created a list of distinct keys that were seen in all proteins for each dataset. This list served as the foundation for our features.  Remarkably, the number of features or unique keys for each dataset exceeds one million.

There are eighteen unique combinations of helix, strand, and coil arrangements that each feature correlates to. As a result, each protein is represented as a two-dimensional matrix in which secondary structure types are rows and unique keys are columns. The cell values indicate how frequently a given key appears within a given structure type. A protein's related cell value is zero when a key is absent. We aggregated the thousands of 2D matrices within each dataset, which resulted in our input data being three-dimensional (Figure \ref{fig:data_struct}). 

We decided to store these matrices in a sparse format to maximize memory utilization. By applying this method, the size of each 2D matrix was reduced by a factor of 20, improving overall efficiency.

\begin{figure*}[!t]
\centering
\includegraphics[trim=0 200 0 140, clip, width=\textwidth]{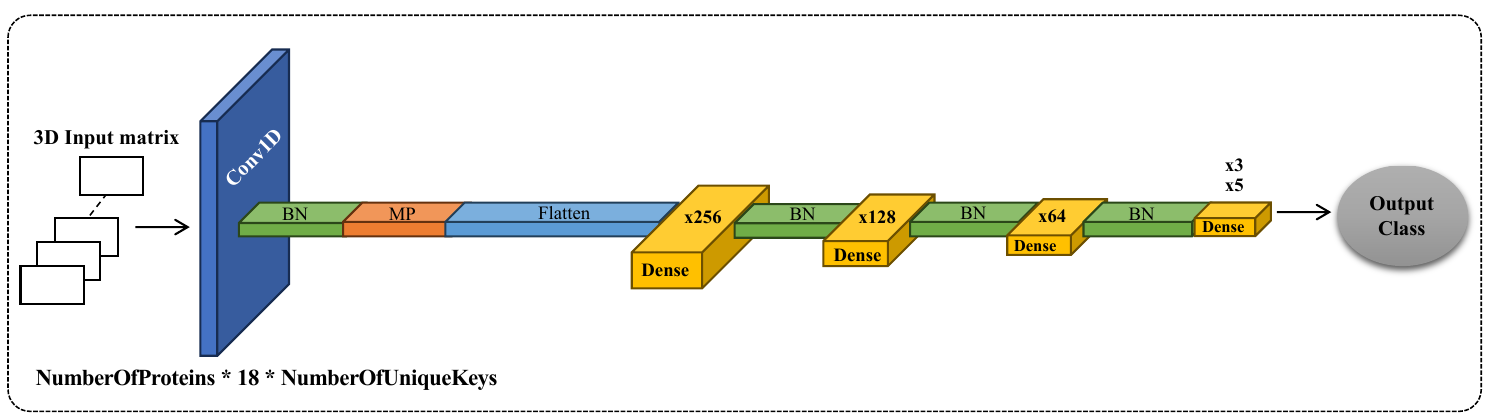}
\caption{Architecture of the Convolutional Neural Network (CNN) used for protein classification. The 3D input matrix, representing the proteins with dimensions NumberOfProteins $\times$ 18 $\times$ NumberOfUniqueKeys, is processed through a Conv1D layer followed by a batch normalization (BN) layer and a max-pooling (MP) layer. The resulting output is then flattened and passed through a series of dense layers interspersed with batch normalization layers. The final dense layer outputs the predicted class.}
\label{fig:cnn_architecture}
\end{figure*}

\subsection{Input Data Handling}

Managing and preparing protein structure data for analysis using machine learning is essential to guarantee accurate and efficient computational research. This section outlines the procedure for efficiently generating, organizing, and processing data batches.

Initially, the label data is obtained from an external CSV file, which is referred to as "sample details." Then, we employed our customizable data generator function, often known as "data generator," which is a keystone of our data processing pipeline. 80\% of the dataset was used for training, 10\% for validation, and 10\% for testing. The dataset directories 'train, validation, and test' include file paths and labels for each class. These are systematically aggregated to make batch generation easier in the future.

The data generation function starts a shuffling operation in every epoch to improve model generalization and add uncertainty. This unpredictability reduces the overfitting risk and guarantees the strength of the trained model. Following shuffling, data batches of a predefined size are produced to enable iterative training of machine learning models.

Each batch undergoes preprocessing to transform the sparse matrix representations of protein structure data into dense matrices of type np.float32. This translation improves computing performance and guarantees interoperability with machine learning techniques, facilitating seamless model training and evaluation.

The limitations on memory presented considerable difficulties during the process of data loading and training, particularly due to the large amount of data involved. Memory overflows were frequently encountered even with the use of HPC (high-performance computing) resources. We effectively optimized memory usage by dynamically loading and processing data in smaller chunks by employing Data generators. This strategy effectively mitigated the risk of memory overflow and guaranteed seamless training for the model.

Ultimately, the data handling process concludes by methodically producing data batches. These batches consist of input data matrices and their matching labels. They play a crucial role in the iterative process of training, validating, and testing machine learning models. The methodical approach to data management and preparation establishes a solid foundation for training and assessing machine learning models. This method ensures the accuracy, effectiveness, and dependability of computational studies of protein structure analysis.

\subsection{Neural Network Model Training and Evaluation}

\begin{figure*}[t!]
\centering
\includegraphics[trim=150 0 150 10, clip, width=0.8\textwidth]{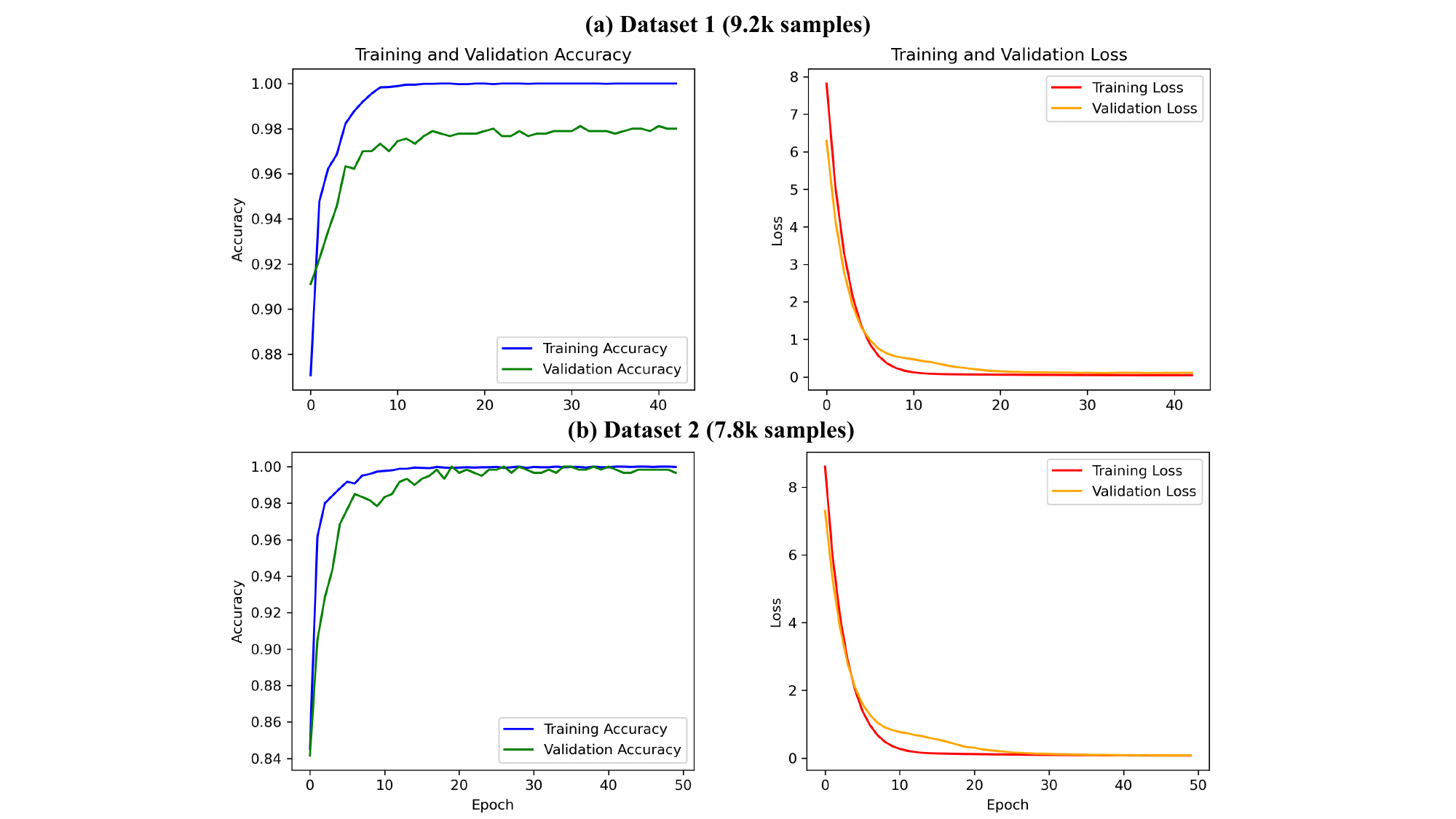}
\caption{Training and validation performance metrics for the datasets with secondary structure information. \textbf{(a)} depicts the training and validation accuracy and loss for Dataset 1, respectively. \textbf{(b)} shows the corresponding plots for Dataset 2.}
\label{fig:acc_loss}
\end{figure*}

To use machine learning to predict protein structures, models must be carefully trained and tested. This section discusses the neural network model's architecture, training procedure, performance evaluation, and overfitting prevention methods.

Figure \ref{fig:cnn_architecture} shows the model's structure, which consists of convolutional, pooling, flattening, and dense layers built with the Keras Sequential API. The model uses convolutional layers with rectified linear unit (ReLU) activation, batch normalization, max-pooling, and dropout regularization.  The last layer uses a softmax activation function to generate probabilistic predictions for various output classes matching the categorical protein structure labels.

Following the architectural specification, the model is compiled using a categorical cross-entropy loss function and the Adam optimizer, with a learning rate set to 0.001. To enhance the model’s generalization ability and prevent overfitting, several techniques were implemented:

\begin{itemize}
    \item Dropout layers were included at many points across the network using a 30\% dropout rate. This method randomly deactivates a fraction of the neurons throughout each training iteration, reducing the reliance on any one feature and increasing robustness. This corresponds with approaches for improving neural network robustness against different perturbations, as reported in recent studies \cite{banitaba2024late}.

    \item The dense layers were subjected to L2 regularization, with a special emphasis on the final layers. This regularization strategy imposes a penalty on big weights, hence preventing the model from becoming excessively intricate and overfitting the training data.

    \item The early stopping strategy was used to monitor the validation loss and stop training after ten consecutive epochs of no progress. This practice helps to prevent overfitting and retains the model's ability to generalize to novel data.

    \item The learning rate was dynamically adjusted during training by applying the learning rate reduction on plateau (ReduceLROnPlateau). Upon observing that the validation loss remained constant for two consecutive epochs, the learning rate was decreased by a factor of 0.2, ensuring that it did not fall below the minimal threshold of 0.0001. This modification enables the model to achieve convergence more efficiently by gradually reducing the size of the steps it takes as it gets closer to the ideal solution.

    \item  Model checkpointing was utilized to preserve the optimal model, determined by the validation loss, guaranteeing that the final model used for evaluation achieved the highest level of generalization performance.
\end{itemize}

Training entails the repetitive process of optimizing model parameters by utilizing batches of protein structure data supplied by the previously mentioned data generator function. The training method comprises 50 epochs, with each epoch consisting of a specific number of steps determined by precomputed steps-per-epoch for the training dataset. Model performance is assessed concurrently on a distinct validation dataset to evaluate its ability to generalize and prevent overfitting. After training is complete, the model is tested on a different dataset to evaluate its performance using the evaluate function. This assessment provides key metrics such as test loss and accuracy, which reflect how well the model can predict outcomes for new, unseen protein structure data. Additionally, predictions are made for each category in the test dataset, enabling the creation of a confusion matrix to further analyze the model’s accuracy and effectiveness in classifying individual groups.

\section{Results}

We tested our SSE-TSR approach on two different protein datasets, comprising 9.2K samples (Dataset 1) and 7.8K samples (Dataset 2), to assess its effectiveness. We evaluated the performance of our method, which integrates secondary structure (SS) information, and contrasted it with the baseline performance, denoted as the SSE-TSR and TSR datasets, respectively. The findings are thoroughly presented in Table 1.

\begin{table}[h!]
\caption{Performance metrics for the models trained on SSE-TSR data and TSR data.\label{tab:table1}}
\centering
\begin{tabularx}{\columnwidth}{|X|c|c|c|c|}
\hline
\textbf{Model} & \textbf{Accuracy} & \textbf{Precision} & \textbf{Recall} & \textbf{F1-Score}\\
\hline
TSR Dataset 1	& 96.0\% & 96.3\% & 95.8\% & 96.0\%\\
\hline
\textbf{SSE-TSR Dataset 1} & 98.3\% & 98.5\% & 98.3\% & 98.4\%\\
\hline
TSR Dataset 2 & 99.4\% & 99.1\% & 98.0\% & 98.6\%\\
\hline
\textbf{SSE-TSR Dataset 2} & 99.5\% & 99.5\% & 99.1\% & 98.8\%\\
\hline
\end{tabularx}
\end{table}

Based on the data shown in Table 1, it is evident that the inclusion of secondary structure information has a substantial positive impact on the performance measures for Dataset 1. More precisely, the inclusion of SS information resulted in an increase in accuracy from 96.0\% to 98.3\%, precision from 96.3\% to 98.5\%, recall from 95.8\% to 98.3\%, and F1-score from 96.0\% to 98.4\%. This exemplifies the augmented significance of secondary structure data in improving the model's capacity to appropriately classify protein structures.

For Dataset 2, the performance measures were already extraordinarily high, even without SS information. However, when SS information was included, only small gains were noticed. The accuracy improved from 99.4\% to 99.5\%, precision went from 99.1\% to 99.5\%, recall increased from 98.0\% to 99.1\%, and the F1-score advanced from 98.6\% to 98.8\%. The minor yet noteworthy improvements demonstrate that the TSR-based approach consistently achieves excellent results for this dataset, regardless of the presence or absence of supplementary SS data. The utilization of TSR keys alone yielded a level of accuracy that was already close to the highest attainable, indicating that Dataset 2 is intrinsically more amenable to precise classification. Although the results show that incorporating SS information can still provide little additional benefits, the potential for further development is restricted because the baseline measures are already quite high.

Figure \ref{fig:acc_loss}a displays the accuracy and loss curves for the training and validation of Dataset 1, including SS information. The training accuracy curve has a steep incline and eventually reaches a perfect score of 100\% (1.0), whereas the validation accuracy plateaus at approximately 98\%. The loss curves exhibit a steep decline, with both the training and validation loss converging towards zero, suggesting successful learning and a minimal likelihood of overfitting.

Dataset 2's accuracy and loss contours are illustrated in Figure \ref{fig:acc_loss}b, which employs SS information. The training accuracy curve swiftly approaches a value of approximately 100\% (1.0), while the validation accuracy reaches a plateau at approximately 99\%. Moreover, the loss profiles of this dataset show a fast decrease; the training and validation loss approaches zero, therefore supporting the extraordinary performance of the model.

The confusion matrices provide a detailed breakdown of the model’s performance across different classes. Both confusion matrices (Figures 5a and 5b) for the datasets with secondary structure information demonstrate high recall for all classes, highlighting the model’s robust performance in correctly identifying the true positive rates across these categories.

\begin{figure}[H]
\centering
\includegraphics[trim=0 90 0 90, clip, width=3.5in]{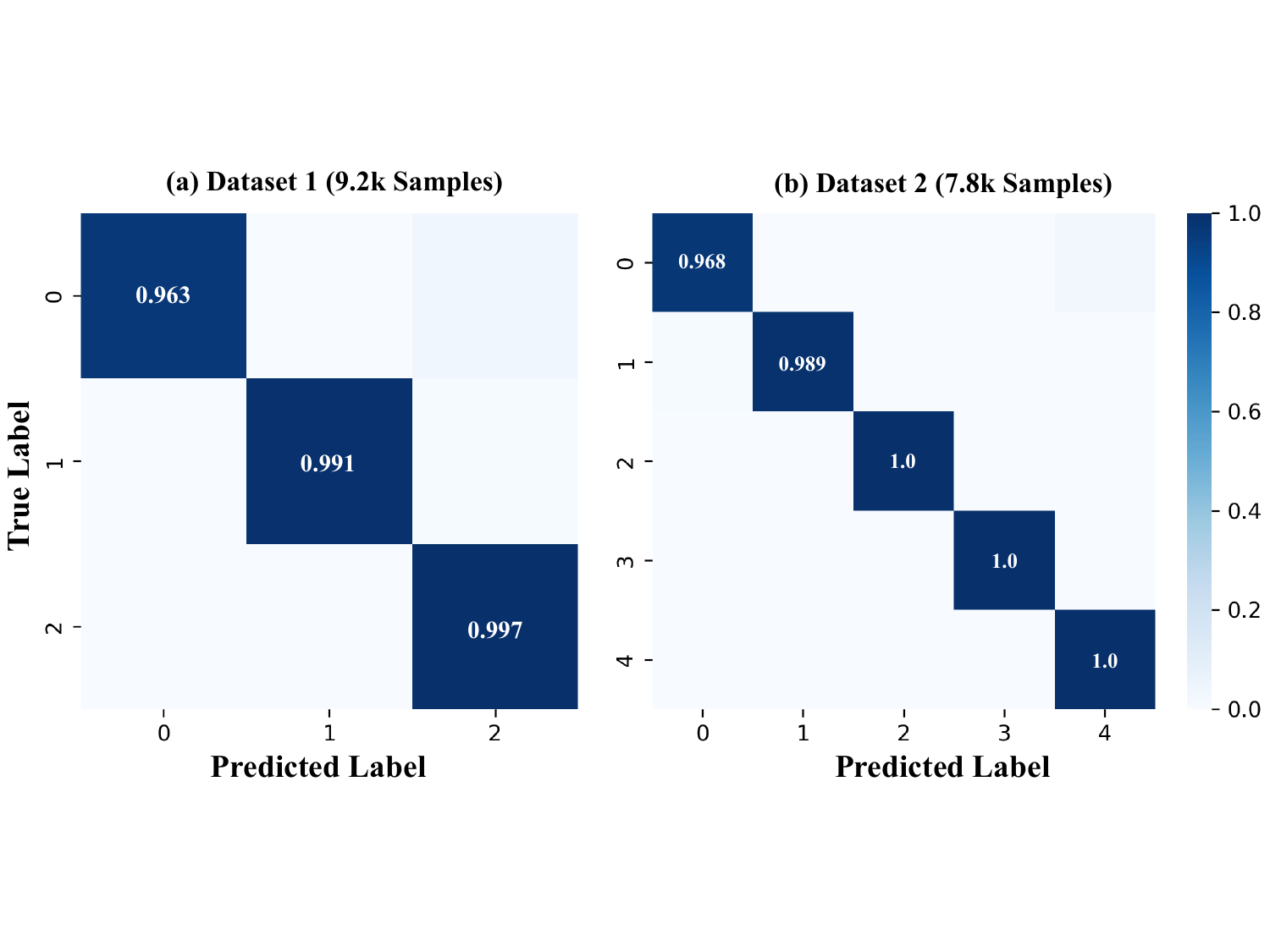}
\caption{Confusion matrices for the models trained on datasets with secondary structure information. \textbf{(a)} Dataset 1 and \textbf{(b)} Dataset 2. Each matrix compares the true labels against the predicted labels, providing a visual representation of the classification accuracy across different classes.}
\label{LP}
\end{figure}

% ==================
% # IV. CONCLUSION #
% ==================

\section{Discussion}

Integrating secondary structure information into the TSR-based technique has shown significant advantages in protein structure classification, especially for bigger datasets. The enhanced metrics for Dataset 1 indicate that incorporating secondary structure data increases the model's ability to accurately reflect the intricate structural interactions within proteins, hence increasing its discriminatory power.

The impressive performance exhibited for Dataset 2, even in the absence of secondary structure information, indicates that the SSE-TSR technique may offer only marginal improvements and that the primary TSR keys alone are enough for precise protein categorization in some scenarios. Due to the already high accuracy (99.4\%) of the TSR-based technique, the potential for additional enhancement by using SS information is intrinsically restricted.
Notably, the disparities between the two datasets could elucidate the differing effects of incorporating secondary structural information. The classification of Dataset 1 is based on topology, which specifically refers to the protein's structural arrangement. Therefore, the incorporation of SS information, which is directly pertinent to the structure, enhances the model's performance. Conversely, Dataset 2 is classified according to its function. In this scenario, the correlation between the structural characteristics represented by secondary structure elements and functional classification might be less apparent. This may clarify why the addition of SS information only slightly improved the already outstanding performance of the TSR-based approach for Dataset 2.

In conclusion, our findings demonstrate that the SSE-TSR method is effective for classifying protein structures. Incorporating secondary structure data provides improved precision and dependability. These discoveries create opportunities for the development of advanced bioinformatics tools that can utilize structural data to enhance our comprehension of protein functions and interactions. Subsequent research may investigate the use of this technique on more demanding datasets that exhibit reduced initial classification accuracy.

\section*{Acknowledgments}
The majority of our calculations were executed at LONI. Here, we want to thank the LONI support team, especially thanks to Feng Chen and Oleg Starovoytov.

\bibliography{main}
\bibliographystyle{IEEEtran}

\end{document}